\documentclass{article}
\usepackage{spconf,amsmath,graphicx}
\usepackage{cite}
\usepackage{algorithm}
\usepackage{algorithmic}
\usepackage{amsmath,amssymb,amsfonts}
\usepackage{subcaption}
\usepackage{flushend}

\title{Efficient Leaf Disease Classification and Segmentation using Midpoint Normalization Technique and Attention Mechanism}

\address{Author Affiliation(s)}
\name{Enam Ahmed Taufik$^1$, Antara Firoz Parsa$^1$, Seraj Al Mahmud Mostafa$^2$}
\address{$^1$Department of Computer Science and Engineering, BRAC University, Dhaka, Bangladesh\\
$^2$Department of Information Systems, University of Maryland, Baltimore County, Baltimore, MD, U.S.A.}

\begin{document}

\maketitle

\begin{abstract}
Enhancing plant disease detection from leaf imagery remains a persistent challenge due to scarce labeled data and complex contextual factors. We introduce a transformative two-stage methodology: Mid Point Normalization (MPN) for intelligent image preprocessing, coupled with sophisticated attention mechanisms that dynamically recalibrate feature representations. Our classification pipeline, merging MPN with Squeeze-and-Excitation (SE) blocks, achieves remarkable 93\% accuracy while maintaining exceptional class-wise balance. The perfect F1 score attained for our target class exemplifies attention's power in adaptive feature refinement. For segmentation tasks, we seamlessly integrate identical attention blocks within U-Net architecture using MPN-enhanced inputs, delivering compelling performance gains with 72.44\% Dice score and 58.54\% IoU, substantially outperforming baseline implementations. Beyond superior accuracy metrics, our approach yields computationally efficient, lightweight architectures perfectly suited for real-world computer vision applications.
\end{abstract}


\section{Introduction}
\label{sec:intro}
Image processing plays a crucial role in disease detection \cite{ngugi2021recent, nawaz2022robust, li2023identification, pothen2020detection}, particularly in identifying and localizing diseased regions. In agriculture, effective preprocessing techniques are essential for enhancing data quality, improving feature extraction, and ultimately boosting the performance of classification and segmentation models. Despite significant advances in machine learning, achieving high accuracy with lightweight models remains challenging, especially in resource-constrained environments where computational efficiency is paramount. 
Lightweight transfer learning models, such as MobileNetV2 \cite{sandler2018mobilenetv2} and VGG16 \cite{simonyan2014very}, have shown promise in disease classification tasks by balancing performance with computational cost. However, these models often struggle when applied to images preprocessed with Mid-point Normalization (MPN), likely due to architectural limitations that hinder their ability to capture complex features. Preprocessing methods like Contrast Limited Adaptive Histogram Equalization (CLAHE) \cite{zuiderveld1994contrast} have enhanced image quality and classification accuracy, raising a critical question: \textit{Why do lightweight models underperform when applied to MPN-preprocessed images, and what strategies can be employed to overcome this limitation?} 

We hypothesize that the poor performance stems from lightweight models' inability to effectively process features extracted from MPN-enhanced images. To address this limitation, we propose the SE-ConvNet architecture, which combines a simplified squeeze-and-excitation attention block \cite{hu2018squeeze} with a simple CNN. This architecture dynamically recalibrates feature maps, improving classification accuracy without sacrificing computational efficiency.
For segmentation tasks, models are essential for identifying and localizing diseased regions in agricultural images \cite{mzoughi2023deep, divyanth2023two}. While U-Net \cite{ronneberger2015unet}, a popular segmentation architecture, yielded satisfactory results in our experiments, performance improved significantly when the simplified SE block was integrated into the U-Net CNN layers. SE blocks enhance feature selection by enabling the model to focus on the most relevant regions, thereby improving segmentation accuracy.

This work focuses primarily on classification and segmentation tasks as key applications. Our main research aim is to bridge existing gaps in disease detection by integrating optimized image-processing techniques with attention mechanisms. By incorporating MPN and SE blocks, we aim to outperform current methods in both classification and segmentation, demonstrating the potential of lightweight, high-performance models in real-world disease detection and localization tasks. The primary research objectives are:
\textit{\textbf{1) A custom CNN architecture integrating a simplified SE block.}} This combination enhances feature selection capabilities and improves classification accuracy without compromising computational efficiency. We demonstrate how SE blocks can dynamically recalibrate feature importance in lightweight models.
\textit{\textbf{2) U-Net with simplified SE block integration.}} We evaluate and compare segmentation model performance, particularly U-Net, in localizing diseased regions in plant leaf images, assessing SE blocks' contribution to improved segmentation performance and model robustness in real-world applications.
\textit{\textbf{3) Comparative analysis of image preprocessing techniques.}} We comprehensively analyze and compare state-of-the-art image preprocessing techniques, assessing their impact on lightweight model performance in classification tasks, including MPN and other methods such as CLAHE.

\section{Image Preprocessing}
\begin{figure}
\centering
\subfloat[Bacterial]{\label{subfig:baclf} \includegraphics[width=0.115\textwidth]{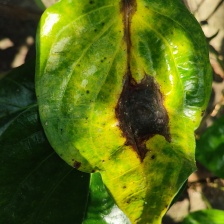}}
\hfill
\subfloat[Dried]{\label{subfig:drdlf} \includegraphics[width=0.115\textwidth]{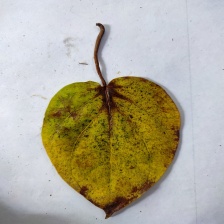}}%
\hfill
\subfloat[Fungal]{\label{subfig:fngllf} \includegraphics[width=0.115\textwidth]{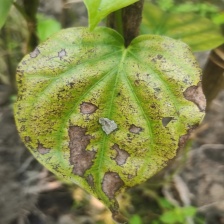}}%
\hfill
\subfloat[Healthy]{\label{subfig:hlthlf} \includegraphics[width=0.115\textwidth]{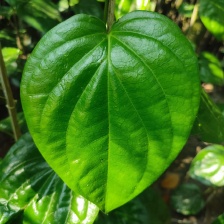}}%
\caption{Betel Leaf Dataset Showing Four Disease Classes (a - d) with 250 Samples Each.}
\label{fig:data-ov}
\end{figure}

\begin{figure}
\centering
\subfloat[Original]{\label{subfig:orig} \includegraphics[width=0.115\textwidth]{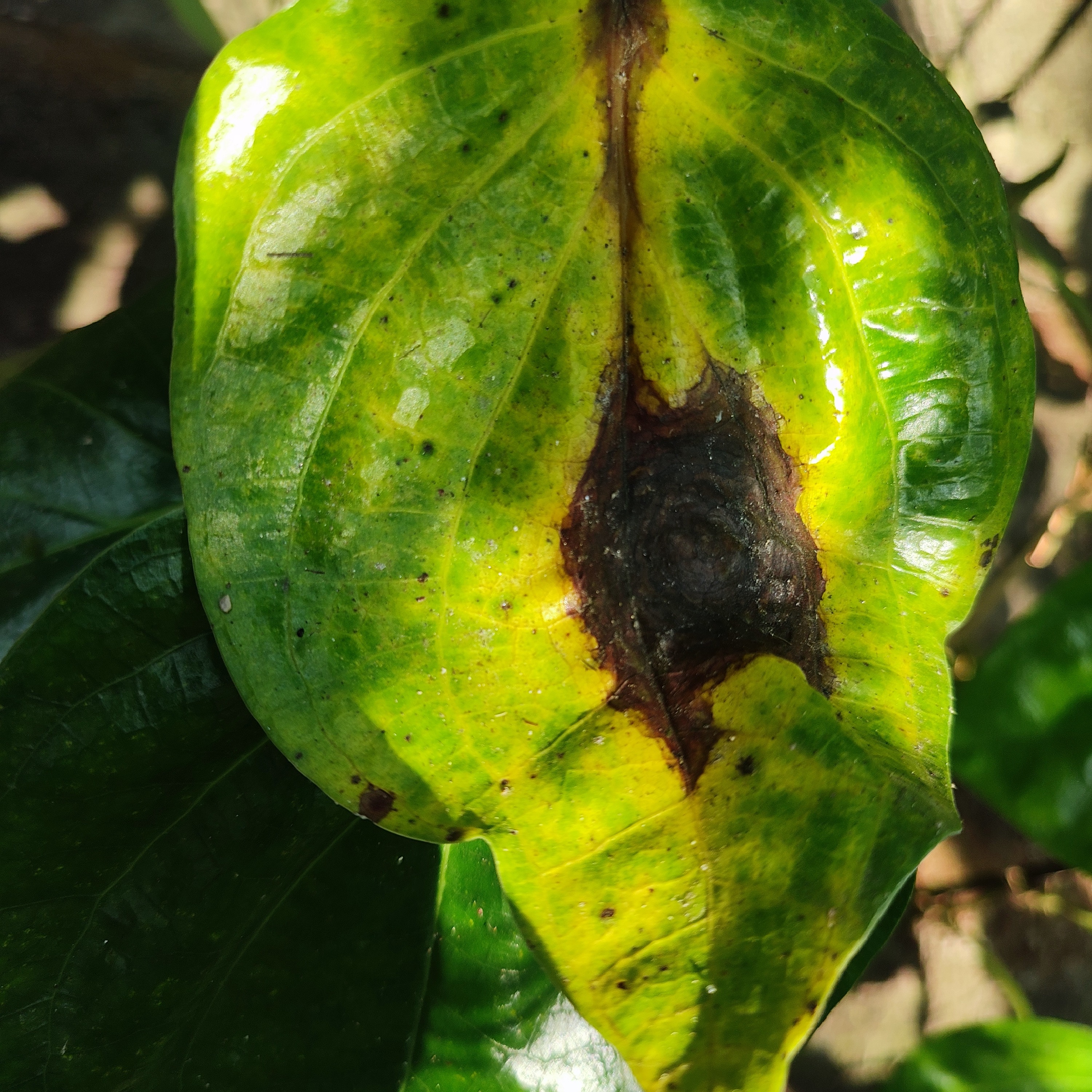}}
\hfill
\subfloat[Resized]{\label{subfig:resi} \includegraphics[width=0.115\textwidth]{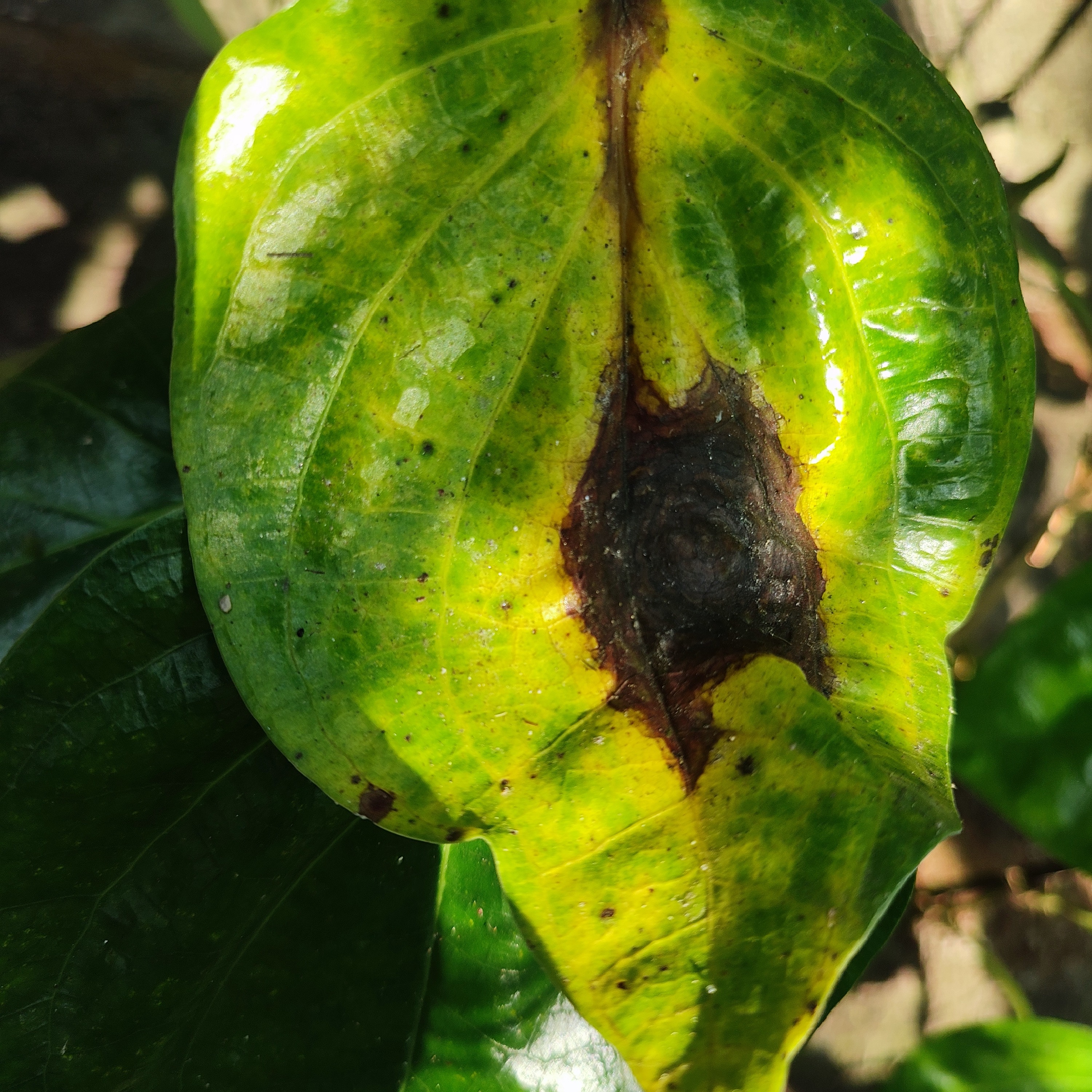}}%
\hfill
\subfloat[CLAHE]{\label{subfig:clahe} \includegraphics[width=0.115\textwidth]{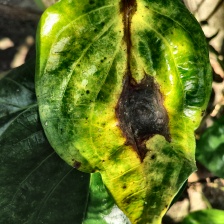}}%
\hfill
\subfloat[MPN]{\label{subfig:mpn} \includegraphics[width=0.115\textwidth]{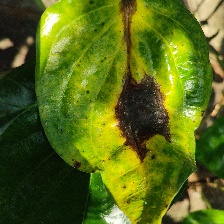}}%
\caption{Comparison of Original and Preprocessed Images Across Various Methods (a - d).}
\label{fig:prep}
\vspace{-1.5em}
\end{figure}

The dataset used in this work comprises images of Betel leaves \cite{rashid2024betel}, categorized into four conditions: healthy, dried, bacterial leaf disease, and fungal brown spot disease (Fig.~\ref{fig:data-ov}). It contains 1,000 images evenly distributed across these categories. For image processing, we primarily employ MPN alongside preprocessing techniques such as resizing, edge detection, and contrast enhancement (Fig.~\ref{fig:prep}). 
MPN scales pixel values relative to a central reference point, typically the mean or median, to balance intensity distribution and enhance contrast. It ensures standardized input by centering values around a midpoint, making features more distinguishable. This technique improves disease detection performance in Betel leaf images by mapping pixel intensities to the range $[-1,1]$ using the formula:
\begin{equation}
\text{MPN} = \tanh\left(\frac{\text{image\_resized}}{127.5} - 1.0\right),
\end{equation}

The normalization process involves two steps: scaling and centering. Scaling maps pixel values (0 to 255) to $[0,2]$ by dividing by 127.5, while centering shifts this range to $[-1,1]$ by subtracting 1.0. This transformation stabilizes gradients during training, preventing vanishing or exploding gradients, and facilitates effective feature learning by maintaining zero-centered distributions. Unlike CLAHE, which enhances local contrast, or Laplacian filtering, which emphasizes edges, MPN directly aligns with neural network requirements, ensuring computational efficiency and standardized input representation.

Prior to MPN, preprocessing includes image resizing, edge detection, and contrast enhancement to highlight disease-relevant features. Images are resized to uniform $225 \times 225$ pixels, ensuring consistent input dimensions while preserving aspect ratio and key visual characteristics. 
Edge detection employs a Laplacian $3 \times 3$ filter, inspired by \cite{mostafa2025gwavenet} that integrated such filters in CNN-like models, to emphasize disease-affected areas by suppressing uniform regions. Contrast enhancement utilizes Contrast Limited Adaptive Histogram Equalization (CLAHE) to improve visibility of affected regions through localized histogram equalization with a contrast limit. While CLAHE effectively enhances subtle pathological features, it introduces inconsistencies due to lighting variations and non-standardized pixel intensities, negatively impacting model stability. In contrast, MPN normalizes inputs and stabilizes training, leading to improved overall model accuracy by preserving relevant disease features while minimizing artifacts and ensuring consistent contrast across images.

\section{Related Works}
Several studies have explored image classification and segmentation for plant disease detection. Femi and Mukunthan \cite{David2024} employed machine learning for betel leaf disease classification using region-of-interest algorithms for segmentation and GLCM for feature extraction, achieving 97\% accuracy with an Extreme Learning Machine classifier. Kumar et al. \cite{Kumar2024} combined machine learning and deep learning, utilizing k-means clustering for disease region extraction and EfficientNet for classification, improving accuracy and computational efficiency. Ford et al. \cite{ford2025joint} introduced a ConvNeXt-based deep learning model for plant segmentation and spray point detection, addressing CNN limitations in unseen conditions. Hou \cite{Hou2021} proposed potato leaf disease segmentation using graph-cut algorithms with entropy-based superpixel analysis for ROI refinement, employing KNN, SVM, ANN, and RF classifiers. Pandiri and Murugan \cite{Pandiri2025} developed ARM-U-Net, an enhanced U-Net with attention gates and residual paths for segmenting fungal pathogen diseases in potato crops, significantly improving segmentation accuracy.

Attention-based models have been widely used for capturing global structures \cite{li2023atten}, contextual information \cite{niloy2021attention}, and understanding image differences \cite{luo2020convolutional}. Attention mechanisms have also been incorporated in preprocessing, as demonstrated by Song et al. \cite{song2023high}, who utilized convolutional block attention modules to extract key information. For segmentation, Li et al. \cite{li2020attention} proposed a nested attention-aware network to extract task-related features, while Wu et al. \cite{wu2023fibonet} introduced a lightweight model with skip connections. However, these models often involve complex architectures leading to high computational costs and larger models, potentially overlooking important information.

Unlike previous studies focusing on handcrafted features, traditional CNNs, or transfer learning, our work integrates SE blocks in CNNs to enhance channel-wise feature representation. Additionally, our segmentation pipeline incorporates tailored U-Net architecture with optimized loss functions and augmentation strategies, ensuring superior disease detection in betel leaves.

\section{Methodology}
\begin{figure}[ht] 
    \centering
    \includegraphics[width=225px]{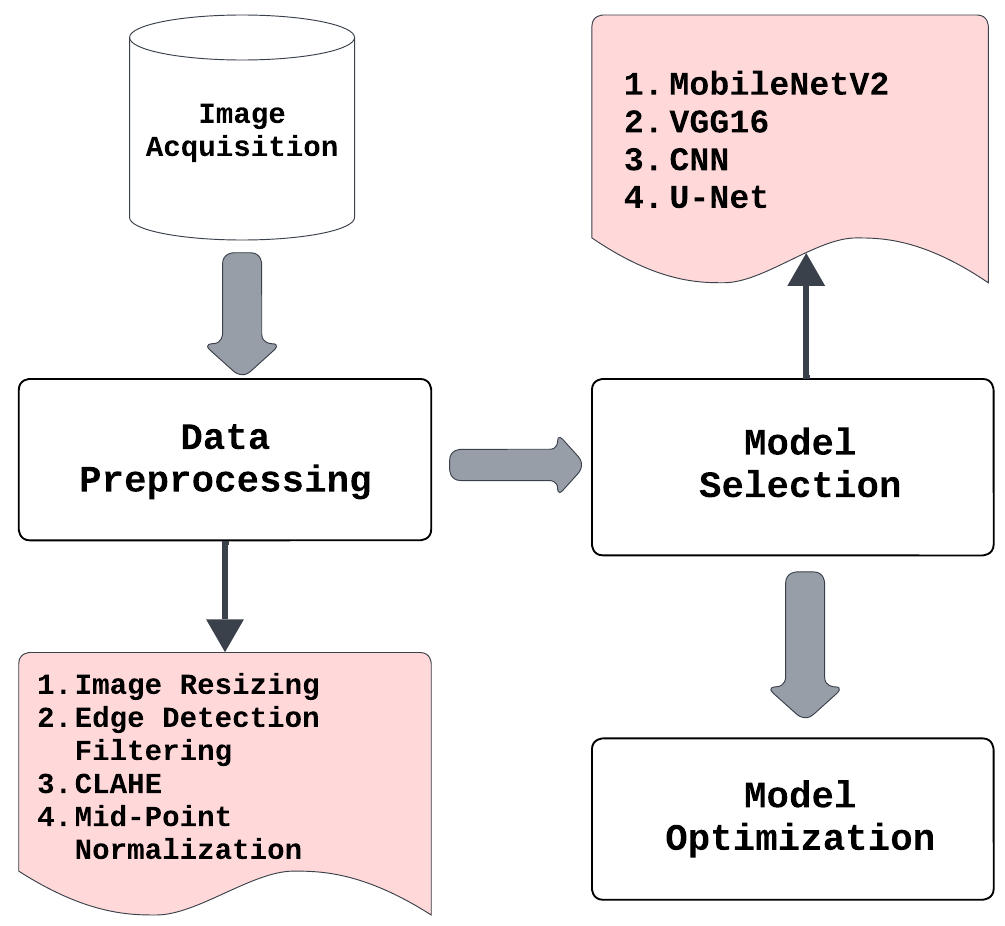} 
    \caption{Overview of the Proposed Workflow.}
    \label{fig:meth}
\end{figure}

This research proposes a systematic approach to betel leaf disease classification and segmentation using deep learning models. We emphasize accuracy, computational efficiency, and interpretability throughout our image processing and model evaluation pipeline. The complete process is illustrated in Fig.~\ref{fig:meth}.

For classification, we employ lightweight models MobileNetV2 \cite{sandler2018mobilenetv2} and VGG16 \cite{simonyan2014very}, alongside our proposed SE-ConvNet architecture. VGG16's 16-layer network with 3×3 convolution filters effectively captures disease patterns through hierarchical feature extraction, while MobileNetV2 employs depthwise separable convolutions, balancing performance with computational efficiency for resource-constrained environments. Additionally, we implement a custom CNN architecture and our novel SE-ConvNet to optimize model performance for betel leaf disease classification. For semantic segmentation, U-Net \cite{ronneberger2015unet} is selected for its superior ability to detect diseased regions in agricultural images. This fully convolutional network employs an encoder-decoder structure that preserves spatial details while capturing semantic features through skip connections that bridge low-level and high-level features. The architecture's symmetric design enables precise localization of disease boundaries, making it particularly well-suited for small agricultural datasets where detailed pixel-level annotation is crucial.
Traditional CNNs treat all feature maps equally, potentially neglecting important channel-specific information. To address this limitation, we integrate Squeeze-and-Excitation (SE) \cite{hu2018squeeze, mostafa2025enhancing} blocks into CNN architectures to incorporate channel-wise attention. 

The SE block processes input tensor $\mathbf{X} \in \mathbb{R}^{H \times W \times C}$ through three sequential operations. First, the \textit{Squeeze} operation applies global average pooling to compress spatial dimensions ($H \times W$) into channel-wise descriptors:
\begin{equation}
z_c = \frac{1}{H \times W} \sum_{i=1}^{H} \sum_{j=1}^{W} \mathbf{X}(i, j, c),
\end{equation}
where $z_c$ represents the global information for channel $c$. Next, the \textit{Excitation} step computes channel-wise attention weights using two fully connected layers with a bottleneck structure:
\begin{equation}
\mathbf{s} = \sigma(\mathbf{W}_2 \, \text{ReLU}(\mathbf{W}_1 \mathbf{z})),
\end{equation}
where $\mathbf{z} = [z_1, z_2, \ldots, z_C]$ is the squeezed feature vector, $\mathbf{W}_1 \in \mathbb{R}^{C/r \times C}$ and $\mathbf{W}_2 \in \mathbb{R}^{C \times C/r}$ are learnable parameters, $r$ is the reduction ratio, and $\sigma$ is the sigmoid activation producing attention weights $\mathbf{s}$. Finally, the \textit{Recalibration} operation applies these attention weights to scale the original feature maps channel-wise:
\begin{equation}
\mathbf{X}'(i, j, c) = \mathbf{s}(c) \cdot \mathbf{X}(i, j, c),
\end{equation}
where $\mathbf{X}'$ is the recalibrated output tensor. SE blocks are integrated after each convolutional block, enhancing feature extraction at all abstraction levels. The complete SE-ConvNet pipeline is detailed in Algorithm~\ref{alg:seconvnet}.

\begin{algorithm}
\caption{Image Classification with SE-ConvNet}
\label{alg:seconvnet}
\begin{algorithmic}[1]
\STATE \textbf{Preprocess Image:} Resize RGB image to \((H, W, C)\)
\STATE \textbf{Normalize Image:} \( \tanh\left(\frac{\text{image\_resized}}{127.5} - 1.0\right) \)
\STATE \textbf{Return:} Mid-Point Normalized image
\STATE \textbf{SE Block: } GAP $\rightarrow$ FC-ReLU $\rightarrow$ FC-Sigmoid $\rightarrow$ Scale
\STATE \textbf{Return:} Recalibrated tensor $X'$ 
\STATE \textbf{Model Architecture:} Conv layers with SE Blocks, pooling, dropout, FC layers, and softmax output.
\end{algorithmic}
\end{algorithm}

Training configuration includes input images resized to 224×224 pixels with three RGB channels and batch size 16, utilizing TensorFlow's \texttt{tf.data} API for efficient data handling. The ReLU activation function introduces non-linearity, while L2 regularization and dropout layers prevent overfitting during training. Training is enhanced using early stopping with patience of 15 epochs, restoring the best weights based on validation loss. A learning rate scheduler, ReduceLROnPlateau, reduces the learning rate by a factor of 0.5 after 15 epochs without validation accuracy improvement, and Adam optimizer with learning rates of 1e-3 for classification models is employed. For segmentation, U-Net utilizes an encoder-decoder architecture with Batch Normalization and 3×3 convolutions for stable training. Conv2DTranspose layers handle upsampling while MaxPooling2D manages downsampling operations. To mitigate overfitting, L2 regularization and dropout are incorporated, while a combined Dice Coefficient and IOU loss function ensures precise segmentation. Data augmentation is performed using a custom \texttt{combined\_generator} function, and training is optimized through callbacks including ReduceLROnPlateau, EarlyStopping, and ModelCheckpoint with Adam optimizer at 1e-4 learning rate.

\section{Results and Discussions}
In our study, we conducted experiments on classification and segmentation tasks. For classification, we evaluated MobileNetV2, VGG16, CNN, and SE-ConvNet under different preprocessing techniques: Edge Detection, CLAHE, and MPN comparing results with raw data to assess their impact on accuracy.

\begin{table}[h]
\centering
\resizebox{\columnwidth}{!}{%
\begin{tabular}{|l|c|c|c|c|c|}
\hline
\textbf{Models} & \textbf{Resized} & \textbf{Edge} & \textbf{CLAHE} & \textbf{MPN} & \textbf{Size (MB)} \\
\hline
MobileNetV2 & 0.37 & 0.23 & 0.76 & 0.239 & 30.41 \\
\hline
VGG16 & 0.81 & 0.77 & 0.83 & 0.24 & 64.92 \\
\hline
CNN & 0.86 & 0.79 & 0.86 & 0.91 & 99.46 \\
\hline
SE-ConvNet & 0.91 & 0.83 & 0.87 & 0.93 & 28.29 \\
\hline
\end{tabular}%
}
\caption{Classification Accuracies of Various Image Pre-processing Techniques.}
\label{tab:preprocessing_accuracies}
\end{table}

Classification results (Table~\ref{tab:preprocessing_accuracies}) showed MobileNetV2 performing the lowest, peaking at 76\% with CLAHE, while VGG16 improved to 83\%. The CNN with MPN achieved 91\%, and SE-ConvNet outperformed all models, reaching 93\%, highlighting its effectiveness alongside advanced preprocessing.

\begin{table}[h]
\centering
\resizebox{\columnwidth}{!}{
\begin{tabular}{|c|c|c|c|}
\hline
\textbf{Classes}                & \textbf{Precision} & \textbf{Recall} & \textbf{F1-Score} \\ \hline
Bacterial Leaf Disease        & 0.90               & 0.88            & 0.89                          \\ \hline
Fungal Brown Spot Disease     & 0.97               & 0.99            & 0.98                         \\ \hline
Dried Leaf                    & 1.00               & 1.00            & 1.00                      \\ \hline
Healthy Leaf                  & 0.86               & 0.88            & 0.87                           \\ \hline
\multicolumn{4}{|c|}{\textbf{Accuracy:} 0.93}  \\ \hline
\textbf{Macro Average}        & 0.93               & 0.93            & 0.93                          \\ \hline
\end{tabular}}
\caption{Performance Metrics of the Best Performing Model.}
\label{tab:best_model_metrics}
\end{table}

\begin{figure}[ht] 
    \centering
    \includegraphics[width=225px]{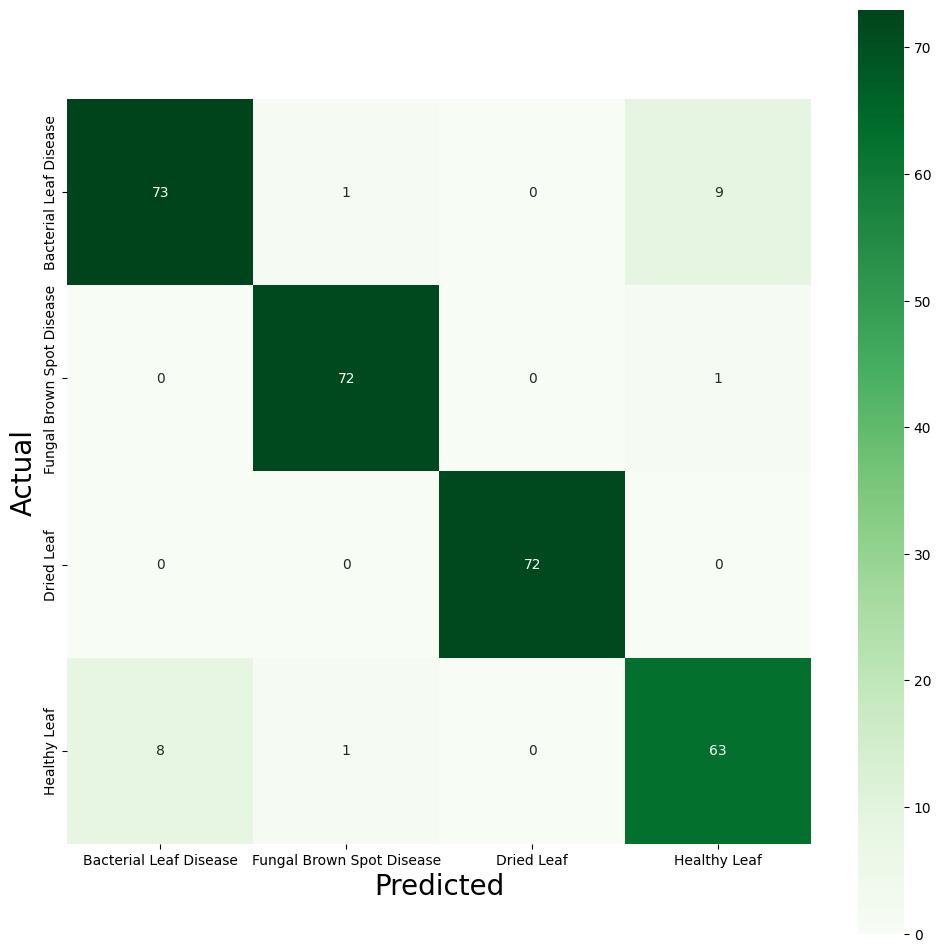} 
    \caption{Confusion Matrix from the SE-ConvNet.}
    \label{fig:cf}
\end{figure}

SE-ConvNet (Table~\ref{tab:best_model_metrics}) excelled in classifying four leaf disease categories: Bacterial Leaf Disease, Fungal Brown Spot Disease, Dried Leaf, and Healthy Leaf, achieving 93\% accuracy with average precision, recall, and F1-score of 0.93. The model performed exceptionally well in detecting Dried Leaf (precision, recall, F1 = 1.00) and Fungal Brown Spot (0.97, 0.99, 0.98). However, Bacterial Leaf (0.90, 0.88, 0.89) and Healthy Leaf (0.86, 0.88, 0.87) showed slight misclassification due to feature similarities, as shown in the confusion matrix (Fig.~\ref{fig:cf}). Notably, 8 Healthy Leaves were misclassified as Bacterial Leaf Disease and 9 vice versa, indicating overlapping features.

\begin{table}[h]
\centering
\begin{tabular}{|l|c|c|}
\hline
\textbf{Models} & \textbf{IoU (\%)} & \textbf{Dice (\%)} \\ 
\hline
U-Net  & 55.755            & 70.791 \\ 
\hline
U-Net + Attention & 58.536 & 72.444 \\ 
\hline
\end{tabular}
\caption{Performance Metrics Comparison Between U-Net and U-Net with Attention Mechanism.}
\label{tab:unet_comparison}
\end{table}

For segmentation (Table~\ref{tab:unet_comparison}), U-Net achieved 55.75\% IoU and 70.79\% Dice score. The U-Net with SE block improved to 58.54\% IoU and 72.44\% Dice, demonstrating the effectiveness of attention mechanisms in enhancing feature selection.

\begin{table}[h]
\centering
\resizebox{\columnwidth}{!}{
\begin{tabular}{|l|c|c|c|}
\hline
\textbf{Classes}                & \textbf{Precision} & \textbf{Recall} & \textbf{F1-Score} \\ \hline
Bacterial Leaf Disease        & 0.76               & 0.78            & 0.77                          \\ \hline
Fungal Brown Spot Disease     & 0.89               & 0.99            & 0.93                         \\ \hline
Dried Leaf                    & 1.0              & 0.99            & 0.99                      \\ \hline
Healthy Leaf                  & 0.79               & 0.69            & 0.74                           \\ \hline
\end{tabular}}
\caption{Performance Metrics of Models without MPN-based Image Pre-processing.}
\label{tab:pre-rec-f1}
\end{table}

\begin{figure}[h]
\centering
\subfloat[U-Net]{\label{subfig:unet} \includegraphics[width=0.235\textwidth]{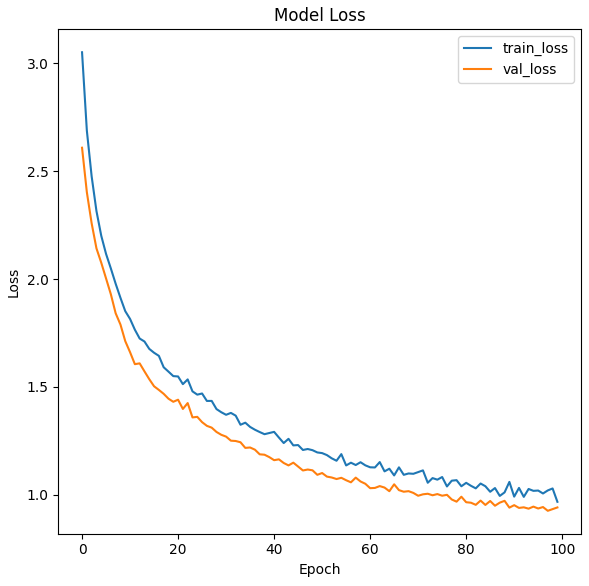}}
\hfill
\subfloat[U-Net with SE block]{\label{subfig:unet-se} \includegraphics[width=0.235\textwidth]{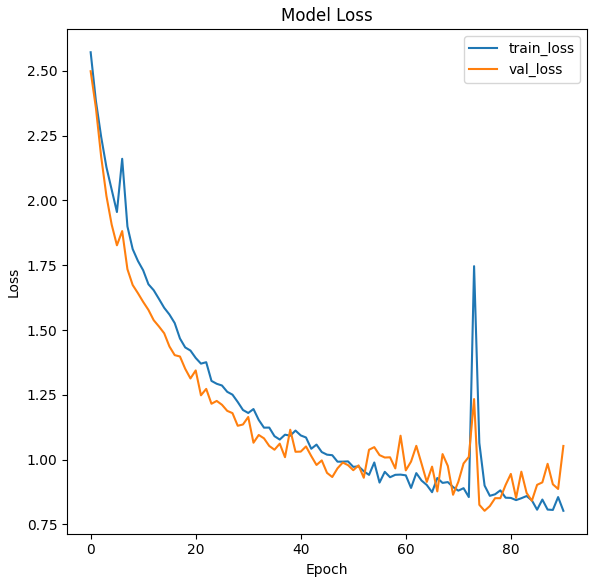}}%
\caption{Comparison of Losses between U-Net and U-Net with SE Block.}
\label{fig:unets}
\end{figure}

\textbf{Ablation Study.} Evaluating MPN and SE-blocks (Tables~\ref{tab:pre-rec-f1} and \ref{tab:no-mpn-unet}), we observed notable improvements. MPN boosted Bacterial Leaf Disease precision (0.76→0.90) and F1-score (0.77→0.89), while Healthy Leaf recall increased from 0.69 to 0.88 (F1: 0.74→0.87). For segmentation, U-Net without MPN yielded 55.75\% IoU and 70.79\% Dice, while adding MPN and attention mechanisms improved IoU to 58.54\% and Dice to 72.44\%, reinforcing MPN's role in spatial detail preservation (Fig.~\ref{fig:unets}).

\begin{table}[ht]
\centering
\begin{tabular}{|l|c|c|}
\hline
\textbf{Models} & \textbf{IoU (\%)} & \textbf{Dice (\%)} \\ 
\hline
U-Net  & 55.75            & 70.79 \\ 
\hline
U-Net with MPN \& Attention & 58.536 & 72.44 \\ 
\hline
\end{tabular}
\caption{Performance Comparison: Base U-Net vs. U-Net with MPN and SE Attention.}
\label{tab:no-mpn-unet}
\end{table}

\begin{figure}[h]
\centering
\subfloat[Original Image]{\label{subfig:grndtrth} \includegraphics[width=0.235\textwidth]{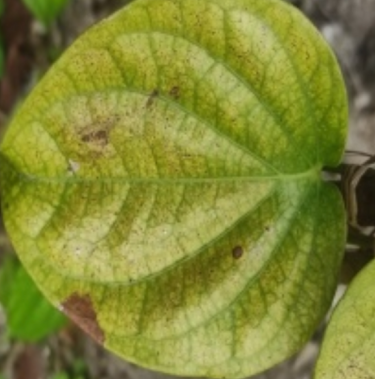}}
\hfill
\subfloat[GradCam Colormap]{\label{subfig:gradcam} \includegraphics[width=0.235\textwidth]{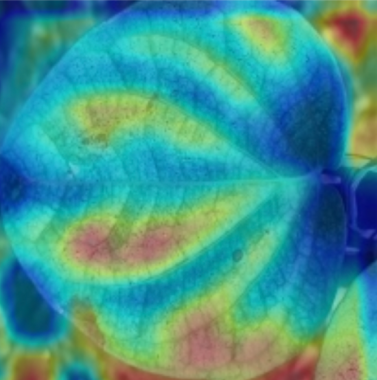}}%
\caption{Feature Visualization using GradCAM++.}
\label{fig:grnd-grdcm}
\end{figure}

\textbf{Discussion.} Our study highlights that preprocessing must be dataset-specific, with our SE-ConvNet architecture achieving 93\% classification accuracy, significantly outperforming MobileNetV2 and VGG16 by 17\% and 10\% respectively. Misclassifications between Healthy and Bacterial Leaf images stemmed from feature inconsistencies, with some bacterial symptoms appearing only at leaf edges, underscoring the need for high-quality labeled datasets. GradCAM++ (Fig.~\ref{fig:grnd-grdcm}) confirms the model's focus on disease regions, enhancing interpretability while demonstrating attention mechanism effectiveness. Additionally, our model is 7.5\% smaller than MobileNetV2 and 56.5\% smaller than VGG16, achieving superior performance with computational efficiency. For segmentation, U-Net with SE blocks improved IoU to 58.54\% and Dice to 72.44\%, representing substantial gains over baseline. Notably, MPN outperformed CLAHE due to better preservation of local contrast and fine edge details, with ablation studies showing precision improvements from 0.76 to 0.90 for Bacterial Leaf Disease. While our approach incorporates SE blocks for attention-based feature recalibration, it relies on a single attention head, which may limit complex inter-class feature capture. Future research could explore multi-head attention mechanisms and deeper architectures to enhance robustness across varying environmental conditions.

\section{Conclusion}
This study demonstrates the effectiveness of integrating Mid-Point Normalization (MPN) with attention mechanisms for plant disease detection. Our SE-ConvNet achieved 93\% classification accuracy, outperforming traditional lightweight models by significant margins while maintaining computational efficiency with a compact architecture. The integration of SE blocks enhanced both classification and segmentation tasks, with U-Net achieving 72.44\% Dice score and 58.54\% IoU. The findings confirm that tailored preprocessing combined with attention-based feature recalibration enables lightweight models to achieve superior performance, making them suitable for deployment in resource-constrained agricultural environments and extensible to other domains facing similar computational challenges.

\bibliographystyle{IEEEbib}
\bibliography{refs}
\end{document}